\documentclass[10pt,twocolumn,letterpaper]{article}

\usepackage[pagenumbers]{cvpr} % To force page numbers, e.g. for an arXiv version

\usepackage{booktabs, multirow, makecell, siunitx, threeparttable}
\usepackage[table]{xcolor}
\usepackage{graphicx}
\usepackage{amsmath,amssymb,mathtools}
\usepackage{booktabs}
\usepackage{microtype}
\definecolor{bestcell}{RGB}{225,240,255} 
\usepackage{xcolor}
\definecolor{bestblue}{RGB}{0,70,180}
\usepackage[normalem]{ulem}
\usepackage{booktabs}
\usepackage{multirow}
\usepackage{siunitx}
\usepackage{xcolor}
\usepackage{tikz}
\usepackage{pgfplots}
\pgfplotsset{compat=1.18}

\definecolor{cvprblue}{rgb}{0.21,0.49,0.74}
\usepackage[pagebackref,breaklinks,colorlinks,allcolors=cvprblue]{hyperref}

\definecolor{cvprblue}{rgb}{0.21,0.49,0.74}
\hypersetup{pagebackref=true,breaklinks=true,colorlinks=true,allcolors=cvprblue}

\usepackage[capitalize,noabbrev]{cleveref}

\def\paperID{18485} % *** Enter the Paper ID here
\def\confName{CVPR}
\def\confYear{2026}

\title{Temporal Realism Evaluation of Generated Videos Using\\ Compressed-Domain Motion Vectors}

\author{
Mert Onur Cakiroglu\\
Indiana University Bloomington\\
{\tt\small meocakir@iu.edu}
\and
\.Idil Bilge Altun\\
Indiana University Bloomington\\
{\tt\small ialtun@iu.edu}
\and
Zhihe Lu\\
Hamad Bin Khalifa University\\
{\tt\small zlu@hbku.edu.qa}
\and
Mehmet Dalkilic\\
Indiana University Bloomington\\
{\tt\small dalkilic@iu.edu}
\and
Hasan Kurban\\
Hamad Bin Khalifa University\\
{\tt\small hkurban@hbku.edu.qa}
}

\begin{document}

\makeatletter
\g@addto@macro\@maketitle{%
  % --- begin teaser in title page ---
  \vspace{-0.75em}%
  \begin{center}
    \includegraphics[width=\linewidth]{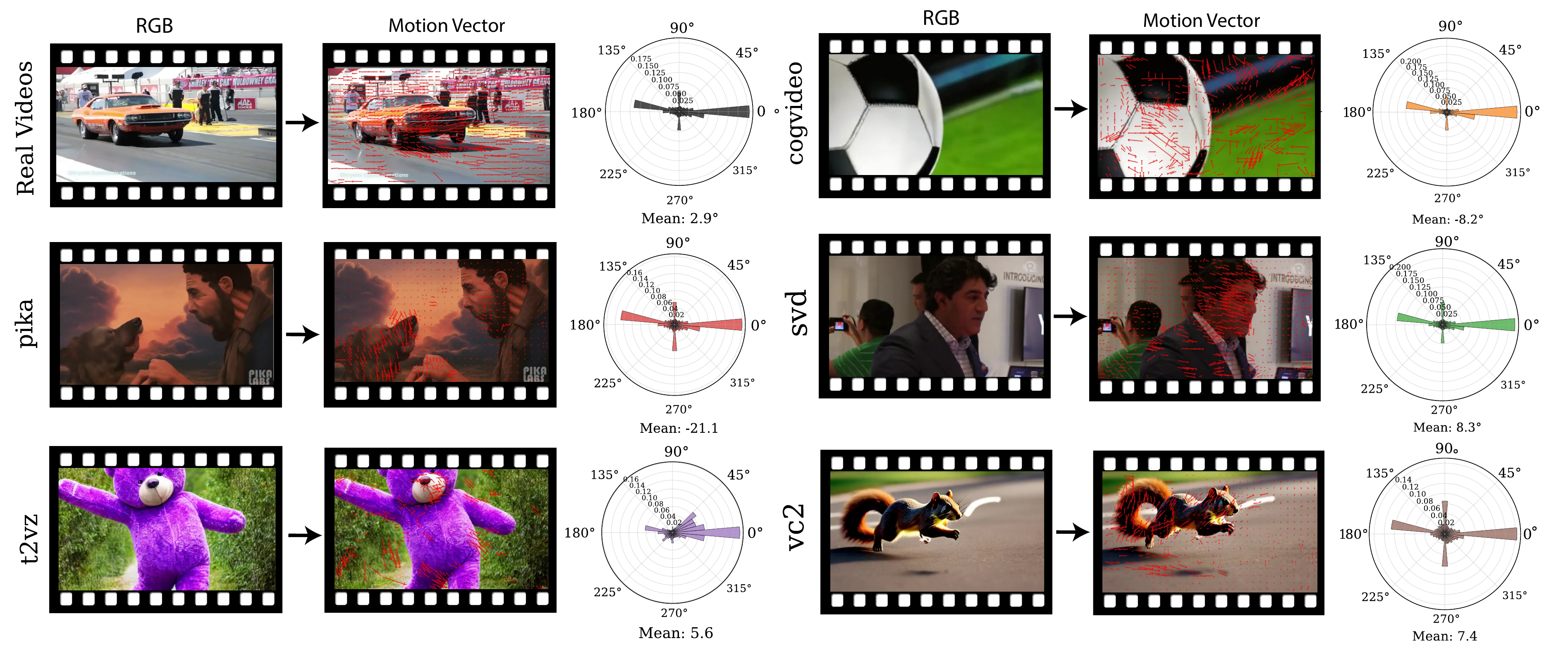}%
    \par\vspace{0.35em}%
    \refstepcounter{figure}% makes this Figure 1
    % Make \label work even without \caption:
    \edef\@currentlabel{\thefigure}\label{fig:direction-distribution}%
    {\footnotesize\bfseries Figure \thefigure.\ %
    Class-conditional directional distributions of motion vectors. Each row corresponds to a dataset or generation model (Real, CogVideo, SVD, Pika, T2VZ, VC2), with representative RGB frames and their associated motion vector visualizations. Each rose diagram depicts the empirical angular density $f_c(\theta)$,
    illustrating the dominant motion orientations per class.}%
  \end{center}
  \vspace{0.5em}%

}
\makeatother
\maketitle

%\begin{abstract}
%Evaluating the temporal fidelity of generative video models remains challenging, as existing metrics largely emphasize spatial quality while overlooking motion realism.  We propose a scalable, model-agnostic framework that measures temporal consistency using motion vectors (MVs) extracted directly from compressed video streams.  Specifically, we repurpose codec-generated MVs (e.g., from H.264 and HEVC) as lightweight, resolution-invariant representations of motion dynamics.  To quantify motion realism, we analyze the distributional divergence between real and synthesized MVs via Kullback–Leibler, Jensen–Shannon, and Wasserstein distances.  Beyond observation, we further investigate several fusion strategies, including MV-RGB concatenation, attention-based fusion, joint embedding and motion-aware fusion, to demonstrate how MVs can effectively enhance temporal reasoning in downstream video tasks.  Extensive experiments show that even state-of-the-art generative models exhibit temporal artifacts missed by frame-based metrics, and that MV-augmented inputs consistently improve video classification performance, particularly for convolutional architectures.
%\end{abstract}
\begin{abstract}
Temporal realism remains a central weakness of current generative video models, as most evaluation metrics prioritize spatial appearance and offer limited sensitivity to motion. We introduce a scalable, model-agnostic framework that assesses temporal behavior using motion vectors (MVs) extracted directly from compressed video streams. Codec-generated MVs from standards such as H.264 and HEVC provide lightweight, resolution-consistent descriptors of motion dynamics. We quantify realism by computing Kullback-Leibler, Jensen-Shannon, and Wasserstein divergences between MV statistics of real and generated videos. Experiments on the GenVidBench dataset containing videos from eight state-of-the-art generators reveal systematic discrepancies from real motion: entropy-based divergences rank Pika and SVD as closest to real videos, MV-sum statistics favor VC2 and Text2Video-Zero, and CogVideo shows the largest deviations across both measures. Visualizations of MV fields and class-conditional motion heatmaps further reveal center bias, sparse and piecewise constant flows, and grid-like artifacts that frame-level metrics do not capture. Beyond evaluation, we investigate MV–RGB fusion through channel concatenation, cross-attention, joint embedding, and a motion-aware fusion module. Incorporating MVs improves downstream classification across ResNet, I3D, and TSN backbones, with ResNet-18 and ResNet-34 reaching up to 97.4\% accuracy and I3D achieving 99.0\% accuracy on real-versus-generated discrimination. These findings demonstrate that compressed-domain MVs provide an effective temporal signal for diagnosing motion defects in generative videos and for strengthening temporal reasoning in discriminative models. The implementation is available at: {\small \url{https://github.com/KurbanIntelligenceLab/Motion-Vector-Learning}}
\end{abstract}

\section{Introduction}
Recent breakthroughs in text-to-video and image-to-video generation have enabled the synthesis of high-resolution \cite{melnik2024videodiffusionmodelssurvey}, visually compelling clips that closely resemble real-world footage. These advances, driven by large-scale diffusion models \cite{text2videozero} and transformer architectures \cite{menapace2024snapvideoscaledspatiotemporal} trained on large-scale paired video and text corpora, increasingly capture long-range dependencies through spatiotemporal attention and improved conditioning. Despite substantial progress in per frame fidelity, assessing and ensuring temporal realism, defined as motion that is physically plausible, stable, and coherent over time, remains a central challenge for evaluation and downstream use \cite{singer2022makeavideotexttovideogenerationtextvideo}.

A growing body of work extends image-generation metrics to video, including Fr\'echet Video Distance (FVD), Inception Score (IS), and perceptual measures such as LPIPS \cite{Unterthiner, Martin, Zhang_2018_CVPR}. Kernel-based distances, for example kernel video distance, provide additional signal beyond frame statistics \cite{binkowski}. These approaches primarily emphasize spatial appearance and often fail to capture subtle motion artifacts such as jitter, drift, or repetition that strongly affect perceptual quality and task performance \cite{unterthiner2019accurategenerativemodelsvideo}. Methods that explicitly model motion, such as optical flow, keypoint trajectories, or motion tracking, are informative but computationally expensive and introduce modeling assumptions that may bias evaluation.

Compressed-domain motion vectors (MVs), computed by standard codecs such as H.264 and HEVC during inter-frame prediction, offer a lightweight alternative for characterizing temporal structure. MVs are resolution-consistent, naturally aligned with frame-to-frame motion, and available at negligible cost during decoding. Building on these properties, we show that statistics of MVs provide a complementary signal for analyzing motion realism in generated videos. When combined with RGB colors, they also enhance downstream video classification without requiring dense flow estimation. This perspective connects to the compressed-domain literature in action recognition, for example CoViAR and DMC-Net \cite{wu2018coviar, shou2019dmc}, but is applied here to generative evaluation and practical scalability.

We quantify motion realism by comparing distributions of MV-based statistics between real and synthetic videos using divergence measures including Kullback-Leibler (KL), Jensen-Shannon (JS), and Wasserstein distances (WD). We also investigate several fusion strategies that combine MV information with RGB features to enrich spatial representations with temporal cues. Together, these components form an efficient and model-agnostic framework that exposes temporal inconsistencies in generative video models and improves classification robustness in scenarios where synthetic videos contain subtle motion artifacts that are difficult to detect with frame-centric metrics.

\begin{figure*}[t]
    \centering
    \includegraphics[width=\linewidth]{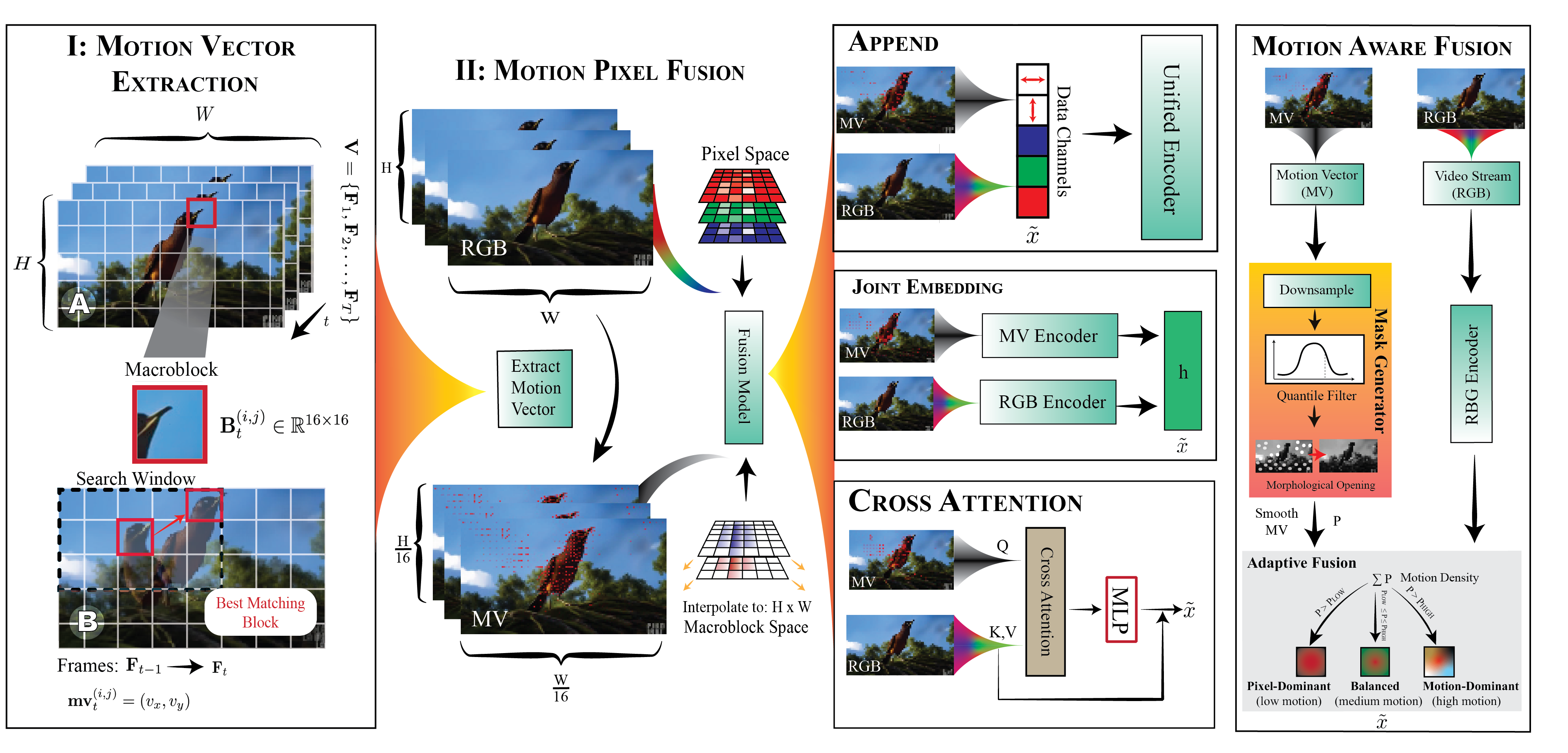}
    \caption{
    Overview of the proposed compressed-domain motion framework. (I) Compressed-domain MVs are obtained from standard video codecs using block matching on macroblocks, for example $16{\times}16$. Frame A illustrates a macroblock, and Frame B shows the search process and the selected best-matching block, which forms the MV between frames. (II) MV fields are spatially aligned with RGB frames to produce consistent motion representations. We consider multi-modal fusion paradigms, including channel concatenation, joint embedding, cross-attention, and motion-aware fusion. In motion-aware fusion, motion masks modulate the contribution of motion features to appearance features, emphasizing informative motion while suppressing noise.
    }
    \label{fig:mv_overview}
\end{figure*}

\noindent\textbf{Contributions.}
Our work makes four contributions:  
(1) We use codec-generated MVs from standard video streams, for example H.264 and HEVC, as an efficient and model-agnostic signal for temporal assessment, providing resolution-consistent motion information at negligible additional cost compared to optical flow;  
(2) We propose a quantitative framework that measures motion realism using divergence metrics including KL, JS, and WD computed between motion statistics of real and generated videos;  
(3) We design and compare several fusion strategies that combine MV information with RGB colors, including channel concatenation, cross-attention, joint embedding, and motion-aware fusion;  
(4) We demonstrate that augmenting RGB with MV channels improves downstream video classification by enriching spatial features with temporal information.

\section{Related Work}
\subsection{Video Generation Models}
The rapid progress in video synthesis has been driven largely by diffusion-based generative models and transformer architectures trained on large-scale video-text corpora. Approaches like VideoCrafter2 \cite{videocrafter2}, Stable Video Diffusion \cite{blattmann2023stable}, and SVD \cite{SVD} employ spatiotemporal attention mechanisms and latent diffusion priors to produce high-resolution videos with strong perceptual fidelity. Contemporary systems differ widely in their data regimes and architectural choices. For instance, SVD and CogVideo rely on the HD-VG130M dataset \cite{hdgv} containing 1280$\times$720, 30 FPS videos, whereas VideoCrafter2, Pika, and Text2Video-Zero are trained on the VidProm dataset \cite{videocrafter2,pika2024,text2videozero}. These systems also vary in their output characteristics: Pika generates 1088$\times$560 videos at 24 FPS; VideoCrafter2 produces 512$\times$320 videos at 10 FPS; Text2Video-Zero synthesizes 512$\times$512 videos at 4 FPS using training-free motion-aware latent codes; SVD outputs 1024$\times$576 videos at 10 FPS via a three-stage training strategy; and CogVideo generates 480$\times$480 videos at 4 FPS with multi-frame-rate hierarchical generation \cite{blattmann2023stable,cogvideo}. Despite these architectural innovations and perceptual improvements, evaluating the temporal realism of synthetic motions remains an unsolved challenge.

\subsection{Video Quality and Temporal Evaluation Metrics}
Most classical video evaluation metrics originate from image generation and primarily target frame-level fidelity rather than motion consistency. Metrics such as FVD and IS \cite{Unterthiner,Martin} do not adequately capture subtle temporal discontinuities that compromise realism. Although computing Fr\'echet Distance on spatiotemporal I3D features \cite{binkowski,Zhang_2018_CVPR} offers improved sensitivity to temporal structure, these methods are computationally expensive and often fail to detect fine-grained inconsistencies crucial for downstream tasks. As diffusion-based models continue to improve in per-frame quality, the lack of reliable temporal realism metrics has become a critical bottleneck for evaluating advances in generative modeling.

\subsection{Video Compression and Motion Representations}
Video compression standards such as AVC/H.264 and HEVC \cite{sullivan2012overview} have long used motion modeling to reduce temporal redundancy. These codecs operate through intra-frame prediction and inter-frame motion compensation, leveraging I-, P-, and B-frame structures to efficiently encode temporal changes \cite{SOLANACIPRES200999}. Following motion prediction, blocks are transformed using DCT, quantized, and entropy-coded using methods such as CAVLC and CABAC \cite{richardson2010h264}. The fundamental reliance on MVs for temporal prediction has made them an abundant, low-cost byproduct of the compressed domain.

\subsection{Motion Vectors in Machine Learning}
Although originally intended for compression, MVs have emerged as powerful temporal features for machine learning. The CPGA network \cite{zhu2024cpga} exploits MVs as coding priors to guide temporal alignment, warping features from neighboring frames while maintaining spatial-temporal coherence for video quality enhancement. Learned video compression models such as MMVC \cite{liu2023mmvc} integrate MVs into block-based predictive modes, mirroring classical codec strategies to improve rate-distortion efficiency. In video-language contexts, MVs support early temporal grounding \cite{fang2023grounding}, enabling efficient linkages between text queries and temporal segments of compressed videos. MVs have also been used in compression-aware super-resolution \cite{wang2023compressionaware} for temporal alignment across frames, improving the fidelity of reconstructed high-resolution videos. Together, these works highlight the versatility of MVs as efficient, information-rich temporal descriptors that bridge traditional video compression and modern machine learning pipelines.

\section{Methods}
\subsection{Motion Vectors}

\begin{figure*}[!htb]
    \centering
    \includegraphics[width=0.99\linewidth]{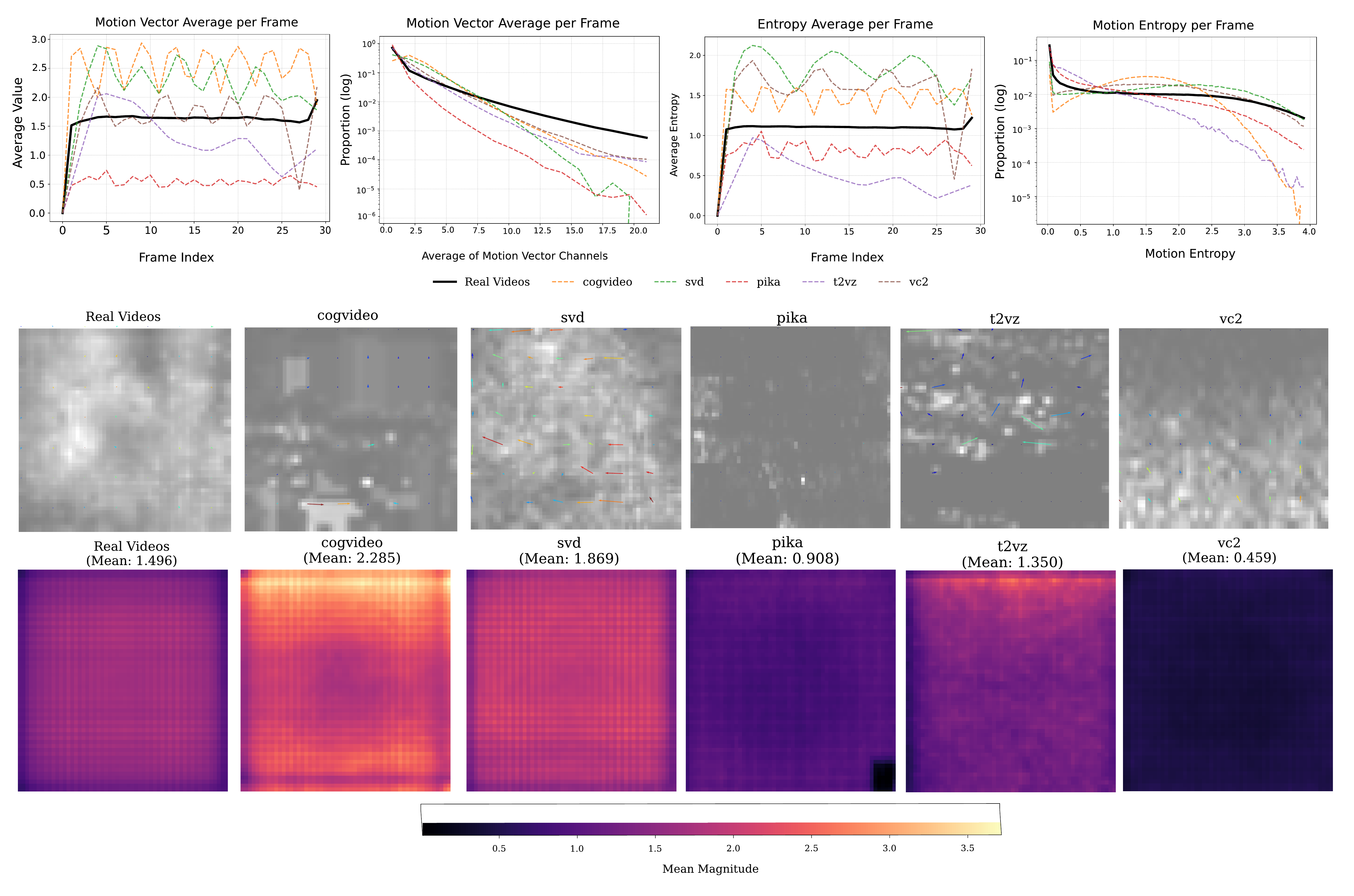}
    \caption{(\textsc{TOP}) Motion statistics of real and generated videos. Videos are temporally interpolated to a common length, after which per-frame averages are computed to visualize how MV sums and motion entropy evolve over time. This view emphasizes temporal dynamics and highlights deviations in frame-to-frame consistency. Temporal ordering is disregarded, and all frames across the dataset are aggregated as independent samples, producing overall distributions of motion vector magnitudes and motion entropy. This representation captures the global statistical alignment of motion features while ignoring sequential context.(\textsc{MIDDLE}) Representative flow fields per class. Arrows indicate the direction of \(V_{t^\star}(u,v)\) and their lengths encode relative magnitude after panel-wise normalization; fields are resized to a common grid and downsampled to a coarse lattice for clarity. More information is given in the Appendix section \ref{subsec:flowfields}. (\textsc{BOTTOM}) Class-conditional motion-magnitude heatmaps. Intensity encodes the expected per-pixel \(\ell_2\) magnitude \(\bar{h}_c(u,v)\), averaged over frames and clips on a \(56\times56\) grid; maps are min–max rescaled to \([0,1]\) for visualization. More information is given in the Appendix section \ref{subsec:heatmap}}

    \label{fig:dist}
\end{figure*}

Videos are sequences of consecutive frames placed in temporal order where each frame consists of a number of pixels. The compression techniques are used to avoid redundancies of both spatial and temporal space. Let a video be characterized by:
\vspace{-0.2em}
\begin{equation}
\mathcal{V} = \{F_1, F_2, \dots, F_T\}, \quad F_T \in \mathbb{R}^{H \times W},
\end{equation}
 where $F$ is the frame in the number of $T$, $H$ and $W$ are the height and width of the frames. Each frame is divided into non-overlapping square regions called macroblocks, typically the size of $16 \times 16$ pixels. %burda başka sizelarda da olabilir sanki kontrol et
Macroblocks in frame $F_t$ are denoted as:
$
\mathcal{B}_t = \{\, B_t^{(i,j)} \mid 
i=1\!:\!\tfrac{H}{M},\;
j=1\!:\!\tfrac{W}{M} \,\},\quad
B_t^{(i,j)}\!\in\!\mathbb{R}^{M\times M},
$
\noindent
where $B$ is macroblock in the spatial position of $(i,j)$. These macroblocks are used as a unit for compression, shown in Figure \ref{fig:mv_overview} (I). The frames are categorized according to their encoding role: $\mathcal{T}_t \in \{\text{I}, \text{P}, \text{B}\}$: 
I-frames (Intra-coded), B-frames (bi-directional predictive-coded) and P-frames (predictive-coded). I-frames are encoded independently by using only spatial information and act as a random access points in group of pictures (GOP); P-frames are encoded by referencing past frames ($\mathcal{F}_{t-k})$ while containing MVs and residuals, and B-frames are compressed with comparison from both past and future frames $(\mathcal{F}_{t-k}, \mathcal{F}_{t+k})$. Let \(F_t\) be the current frame and \(F_{t+k}\) a reference frame with temporal offset \(k\in\mathbb{Z}\).
For macroblock indices \((i,j)\), define the current block \(B_t^{(i,j)}\) and search a discrete window
\(\mathcal{S}\subseteq\{-S,\dots,S\}^2\) in the reference frame to find the best match. With
\vspace{-0.4em}
\begin{equation}
\begin{aligned}
A(u,v) &= B_t^{(i,j)}(u,v), \\
B_k(u,v) &= B_{t+k}^{(i+v_x,\; j+v_y)}(u,v),
\end{aligned}
\label{eq:blockpairs}
\end{equation}
\noindent
The similarity cost (e.g., SAD) is
\begin{equation}
D_{\mathrm{SAD}}(A,B_k)
  = \sum_{u=1}^{M}\sum_{v=1}^{N} 
    \lvert A(u,v)-B_k(u,v)\rvert .
\label{eq:sad}
\end{equation}
\noindent
The MV and its temporal sign are
\begin{equation}
\mathbf{mv}_t^{(i,j)}
=
\arg\min_{(v_x,v_y)\in\mathcal{S}}
D_{\mathrm{SAD}}(A,B_k)
\end{equation}

\begin{equation}
t_{\mathrm{sign}}^{(i,j)}
=
\operatorname{sign}(k)\in\{-1,0,+1\}
\end{equation}

\noindent
where \(t_{\mathrm{sign}}=-1\) indicates a backward reference \((k<0)\), \(+1\) a forward reference \((k>0)\), and \(0\) denotes intra coding (no inter reference). The corresponding residual is
\begin{equation}
R_t^{(i,j)}(u,v) = A(u,v) - B_k^{*}(u,v),
\label{eq:residual}
\end{equation}
\noindent
with \(B_k^{*}\) the block at the minimizing displacement. In intra-coded cases \((t_{\mathrm{sign}}=0)\), \(R_t^{(i,j)}\) is taken with respect to the intra predictor of \(B_t^{(i,j)}\).

\subsubsection{Complexity of Motion Vector Extraction}
In H.264 video compression, each frame of size $H$ and $W$ is divided into macroblocks of size $M \times M$. The number of blocks per frame is given by $N_{\mathrm{MB}}
= \left\lceil \frac{H}{M} \right\rceil
  \left\lceil \frac{W}{M} \right\rceil .
$
During decoding, MVs are already available in the compressed bitstream having been computed for inter-frame prediction in encoding, making extraction linear to the number of macroblocks and frames: Let $T$ denote the number of frames and $N_{MB}$ be the number of macroblocks in each frame. The overall complexity of MV extraction is, therefore, $\mathcal{O}\!\left(T\, N_{\mathrm{MB}}\right).$

\subsection{Feature Fusion}
To standardize input dimensions and aspect ratios across videos, we adopt a two-step preprocessing pipeline. Each video frame is first center-cropped to match the target aspect ratio without distortion. This preserves the spatial layout while eliminating the need for stretching. The cropped frames are then resized to a fixed resolution $H \times W$, forming the spatial standard for all subsequent processing. MVs, extracted from the compressed video stream, are resolution-dependent by nature because they are encoded relative to the original frame dimensions. To align the motion information with the standardized spatial input, we resize the MV macroblocks to the same resolution $H \times W$ using bilinear interpolation:
$
\mathbf{V}_{\text{resized}}
= \operatorname{Resize}(\mathbf{V}_{\text{orig}},\, H,\, W),
$
here, $\mathbf{V}_{\text{orig}} \in \mathbb{R}^{H_0 \times W_0 \times 3}$ represents the original horizontal and vertical MVs and $t_{\text{sign}}$, and $\mathbf{V}_{\text{resized}} \in \mathbb{R}^{H \times W \times 3}$ is the resized MV field. The RGB channels are then concatenated with the MV components $v_x$ and $v_y$ to form a unified feature tensor per frame:
$
\mathbf{x}_{i,j}
= (R_{i,j},\, G_{i,j},\, B_{i,j},\,
   v^x_{i,j},\,
   v^y_{i,j},\, t_{\mathrm{sign}}),
\quad
\mathbf{x} \in \mathbb{R}^{F \times H \times W \times 6}.
$
This six-dimensional per-pixel representation combines appearance and motion cues, enabling the model to learn both spatial and temporal patterns. The use of MVs provides an efficient alternative to optical flow, requiring no additional computation beyond video decoding.

\subsection{Statistical Alignment of Motion Vectors}

\subsubsection{Motion vector sum and entropy}
Let \(v_t:\Omega\to\mathbb{R}^2\) be the MV field on the pixel grid \(\Omega=\{1,\dots,H\}\times\{1,\dots,W\}\) and let
\(m_t(x,y)=\lVert v_t(x,y)\rVert_2\) be its per-pixel magnitude. Define the per-frame motion sum
\begin{equation}
S_t = \sum_{(x,y)\in\Omega} m_t(x,y),
\label{eq:sum}
\end{equation}
\noindent
and the per-frame mean magnitude (used in our implementation)
\begin{equation}
\bar{m}_t = \frac{1}{HW}\sum_{(x,y)\in\Omega} m_t(x,y).
\label{eq:mean}
\end{equation}
For a histogram with \(K\) bins over \(\{m_t(x,y)\}_{(x,y)\in\Omega}\), let \(p_t(k)\) be the empirical bin probability. The per-frame motion entropy is
\begin{equation}
H_t = -\sum_{k=1}^{K} p_t(k)\,\log_2 p_t(k).
\label{eq:entropy}
\end{equation}
For a clip \(i\) with frame index set \(T_i\), the clip-level descriptors are
\begin{equation}
M_i = \frac{1}{|T_i|}\sum_{t\in T_i} \bar m_t,
\qquad
\mathcal{H}_i = \frac{1}{|T_i|}\sum_{t\in T_i} H_t.
\label{eq:stats}
\end{equation}
\noindent
To quantify how well synthetic videos replicate the temporal motion patterns of real-world videos, we analyze the distributions of MV magnitudes and entropies extracted from each video. Let $P$ denote the distribution of MV statistics from real videos and $Q$ from a synthetic video model. We evaluate the divergence between these distributions using the KL, JS, and WD. The KL from $P$ to $Q$, defined as
\vspace{-0.3em}
\begin{equation}
\mathrm{KL}(P\parallel Q)
= \sum_{x\in\mathcal{X}} P(x)\,\log\!\frac{P(x)}{Q(x)},
\label{eq:kl}
\end{equation}
\noindent
measures how much information is lost when $Q$ is used to approximate $P$. Since KL is asymmetric, we compute both $\text{KL}(P \parallel Q)$ and $\text{KL}(Q \parallel P)$. The JS, which is a symmetrized and smoothed version of KL, is given by

\begin{equation}
\mathrm{JS}(P,Q)
= \tfrac{1}{2}\,\mathrm{KL}(P\parallel M)
+ \tfrac{1}{2}\,\mathrm{KL}(Q\parallel M),
\label{eq:js}
\end{equation}
\noindent
where $M = \frac{1}{2}(P + Q)$. JS is bounded and always finite, making it more stable for empirical comparisons. The WD of order one, also known as the Earth Mover’s Distance, is defined as
\begin{equation}
W(P,Q)
= \inf_{\gamma\in\Gamma(P,Q)}
\mathbb{E}_{(x,y)\sim\gamma}\!\left[\|x-y\|\right],
\label{eq:wasserstein}
\end{equation}
\noindent
where $\Gamma(P, Q)$ denotes the set of all joint distributions (couplings) with marginals $P$ and $Q$. WD measures the minimal ``cost'' of transporting mass from $P$ to $Q$, thus capturing how far distributions are in metric space.

\subsection{Model Fusion Approaches}

%\begin{figure*}
%    \centering
%    \includegraphics[width=.75\linewidth]{mvaf_before_after.pdf}
%    \caption{illustration of the intermediate steps of the motion-aware fusion module. The left panel shows an RGB frame from a video sequence, followed by the middle panel displaying the motion flow field of the frame, where motion vectors are extracted from the compressed stream, indicating pixel-wise displacement between consecutive frames. The right panel visualizes the high-magnitude motion mask obtained by thresholding the top 10 \% of motion-vector magnitudes and refining the result with morphological opening. The resulting binary mask highlights regions of dominant motion.}
%    \label{fig:maskgen}
%\end{figure*}

We implement a comprehensive framework for fusing video and MV features through multiple architectural paradigms. Our approach supports both ResNet and Vision Transformer (ViT) backbones, each with specialized temporal processing mechanisms, illustrated in Figure \ref{fig:mv_overview}  (II).

\subsubsection{Backbone Architectures}

\textbf{ResNet-based Processing.} For ResNet backbones, we employ temporal averaging to aggregate frame-level features. Given input video sequences $\mathbf{X} \in \mathbb{R}^{B \times T \times C \times H \times W}$, where $B$ is batch size, $T$ is temporal length, and $C \times H \times W$ represents spatial dimensions, we process each frame independently: $\mathbf{f}_t = \text{ResNet}(\mathbf{x}_t), \quad t \in \{1, 2, \ldots, T\}$. The temporal aggregation is performed through simple averaging: $\mathbf{f}_{\text{final}} = \frac{1}{T} \sum_{t=1}^{T} \mathbf{f}_t$.

\subsubsection{Multi-modal Fusion Strategies}

We implement four distinct fusion strategies for combining video and MV features.\\
\noindent
\textbf{Channel Concatenation.} The simplest fusion approach concatenates video and MV features along the channel dimension:
$ \mathbf{F}_{\text{concat}} = \text{Concat}(\mathbf{F}_{\text{vid}}, \mathbf{F}_{\text{mv}}) \in \mathbb{R}^{B \times (C_{\text{vid}} + C_{\text{mv}}) \times H \times W}
$, where $\mathbf{F}_{\text{vid}} \in \mathbb{R}^{B \times C_{\text{vid}} \times H \times W}$ and $\mathbf{F}_{\text{mv}} \in \mathbb{R}^{B \times C_{\text{mv}} \times H \times W}$ represent video and MV features, respectively.\\ \\
\noindent
\textbf{Cross-Attention.} We implement a learnable cross-attention mechanism that allows video features to attend to MV features. The cross-attention module computes: $\mathbf{Q} = \mathbf{F}_{\text{vid}} \mathbf{W}_q, \quad \mathbf{K} = \mathbf{F}_{\text{mv}} \mathbf{W}_k, \quad \mathbf{V} = \mathbf{F}_{\text{mv}} \mathbf{W}_v$. The attention weights are computed as
$
\mathbf{A} = \text{softmax}\left(\frac{\mathbf{Q}\mathbf{K}^T}{\sqrt{d_k}}\right)
$. The fused representation is obtained through
$\mathbf{F}_{\text{cross}} = \mathbf{A}\mathbf{V}\mathbf{W}_o,$
where $\mathbf{W}_o$ is the output projection matrix.\\ \\
\noindent
\textbf{Joint Embedding.} For the most sophisticated fusion approach, we process video and MV features through separate embedding networks and employ contrastive learning. Given video features $\mathbf{f}_{\text{vid}}$ and MV features $\mathbf{f}_{\text{mv}}$, we compute normalized embeddings:
$
\mathbf{e}_{\text{vid}} = \frac{\mathbf{f}_{\text{vid}}}{\|\mathbf{f}_{\text{vid}}\|_2}, \quad \mathbf{e}_{\text{mv}} = \frac{\mathbf{f}_{\text{mv}}}{\|\mathbf{f}_{\text{mv}}\|_2}.
$
The final classification is performed on the concatenated embeddings: $\mathbf{y} = \text{Classifier}(\text{Concat}(\mathbf{f}_{\text{vid}}, \mathbf{f}_{\text{mv}}))$.\\ \\
\noindent
\textbf{Motion-aware Fusion (MAF).}
At time \(t\), RGB appearance is \(V_t\in\mathbb{R}^{3\times H\times W}\) and codec MVs are \(M_t=\{d_x,d_y,t_{\text{sign}}\}\in\mathbb{R}^{3\times H\times W}\) with \(t_{\text{sign}}\in\{-1,0,+1\}\) for backward, intra, and forward references. Compute the MV magnitude $m = \sqrt{d_x^2 + d_y^2 + 10^{-12}},
$
$
m \leftarrow m \cdot \mathbf{1}[t_{\text{sign}} \neq 0].
$
For each direction \(s\in\{-1,+1\}\), define \(m_s=m\cdot \mathbf{1}[\mathrm{sign}(t_{\text{sign}})=s]\) and a downsampled map \(\hat m_s=D(m_s)\). Introduce a tunable hyperparameter \(q_{\mathrm{mv}}\in(0,1)\) that specifies the quantile level used for mask selection. Let \(q_s=\mathrm{Quantile}(\hat m_s,\,q_{\mathrm{mv}})\) and set
\begin{equation}
\begin{aligned}
P_s &= \mathbf{1}[m_s \ge q_s] \, \mathbf{1}[\operatorname{sign}(t_{\text{sign}})=s].\\
%P_s &= \mathbb{1}\{m_s \ge q_s\}\,\mathbb{1}\{\operatorname{sign}(t_{\text{sign}})=s\},\\
P_{\mathrm{raw}} &= P_{-1}\lor P_{+1},\\
P &= \mathcal{O}(P_{\mathrm{raw}}),
\end{aligned}
\label{eq:ps}
\end{equation}
\noindent
where \(\mathcal{O}(\cdot)\) is a morphological opening with a small structuring element. The framewise motion density is
$
\rho = \frac{1}{HW}\sum_{i,j} P_{ij}.
\label{eq:rho}
$ To reduce sparsity, blend MVs with a local average \(B(\cdot)\)
\begin{equation}
\tilde M = P\odot M_t + (1-P)\odot B(M_t).
\label{eq:blend}
\end{equation}
Let \(A_t=\phi_V(V_t)\) be appearance features and \(U_t=\mathrm{Resize}(\phi_M(\tilde M))\) be MV features resampled to the spatial size of \(A_t\). A spatial gate
\begin{equation}
G = \sigma\!\big(g(U_t)\big)\odot P^{\uparrow}, 
\qquad 
\hat U_t = G\odot U_t,
\label{eq:gate}
\end{equation}
modulates MV contribution, where \(g(\cdot)\) outputs a scalar map, \(\sigma\) is a sigmoid, and \(P^{\uparrow}\) is the upsampled mask. With density cut points \(0<p_{\text{low}}<p_{\text{high}}<1\) set from data, routing follows
\begin{equation}
F_t=
\begin{cases}
A_t, & \rho < p_{\text{low}},\\[2pt]
f_{\text{mid}}\big([A_t,\hat U_t]\big), & p_{\text{low}}\le \rho \le p_{\text{high}},\\[2pt]
f_{\text{high}}\big([A_t,\hat U_t]\big), & \rho > p_{\text{high}},
\end{cases}
\label{eq:F_t}
\end{equation}
where \(\phi_V\), \(\phi_M\), \(g\), \(f_{\text{mid}}\), and \(f_{\text{high}}\) are learned differentiable modules; \(\phi_V\) encodes RGB appearance, \(\phi_M\) encodes motion cues from codec vectors, \(g\) maps motion features to a spatial gate aligned with the mask, \(f_{\text{mid}}\) performs a RGB-biased blend for moderate motion, and \(f_{\text{high}}\) performs a stronger MV-biased blend for dense motion.

\begin{table}[t]
    \centering
    \scriptsize
    \setlength{\tabcolsep}{3pt}
    \renewcommand{\arraystretch}{1.05}

    \caption{\textbf{RGB+MV fusion improves video classification.}
    Metrics in \%. Best result per backbone in \textbf{bold}, second best are \uline{underlined}. 
    Abbrev.: Concat: feature concatenation, JE: joint embedding, 
    CA: cross attention, MAF: motion-aware fusion, 
    Feat.: features, Prm.: parameter count in millions, 
    Prec.: precision, Rec.: recall.}
    \label{tab:rgbmv-singlecol}

    \resizebox{\columnwidth}{!}{%
    \begin{tabular}{@{}l l l S S S S S@{}}
      \toprule
      \textbf{Backbone} & \textbf{Feat.} & \textbf{Fusion} &
      \textbf{Prm. (M)} & \textbf{Acc.} & \textbf{F1} & \textbf{Prec.} & \textbf{Rec.} \\
      \midrule

      % ----------------- ResNet-18 -----------------
      \multirow{5}{*}{ResNet-18}
        & RGB     & --      & 11.18 & 84.6 & 84.3 & 84.8 & 84.2 \\
        & \multirow{4}{*}{RGB+MV}
        & Concat  & 11.19 & 85.4 & 84.8 & 85.7 & 84.7 \\
        &         & \textbf{JE}   & \textbf{22.68} & \textbf{95.2} & \textbf{95.3} & \textbf{95.3} & \textbf{95.3} \\
        &         & CA      & 11.19 & 91.8 & 91.5 & 91.8 & 91.5 \\
        &         & \uline{MAF}     & \uline{11.39} & \uline{94.2} & \uline{94.1} & \uline{94.2} & \uline{94.1} \\
      \midrule

      % ----------------- ResNet-34 -----------------
      \multirow{5}{*}{ResNet-34}
        & RGB     & --      & 21.29 & 84.1 & 83.5 & 84.5 & 83.5 \\
        & \multirow{4}{*}{RGB+MV}
        & Concat  & 21.30 & 84.4 & 83.7 & 84.8 & 83.6 \\
        &         & \uline{JE }     & \uline{42.90} & \uline{96.2} & \uline{96.5} & \uline{96.5} & \uline{96.5} \\
        &         & \textbf{CA}   & \textbf{21.30} & \textbf{97.4} & \textbf{97.5} & \textbf{97.5} & \textbf{97.5} \\
        &         & MAF     & 21.49 & 93.5 & 93.6 & 93.5 & 93.5 \\
      \midrule

      % ----------------- ResNet-50 -----------------
      \multirow{5}{*}{ResNet-50}
        & RGB     & --      & 23.52 & 83.5 & 83.1 & 83.8 & 83.0 \\
        & \multirow{4}{*}{RGB+MV}
        & Concat  & 23.53 & 84.1 & 83.5 & 84.5 & 83.5 \\
        &         & \textbf{JE}   & \textbf{48.13} & \textbf{93.8} & \textbf{93.6} & \textbf{93.8} & \textbf{93.6} \\
        &         & CA      & 23.53 & 87.8 & 87.9 & 88.0 & 87.9 \\
        &         & \uline{MAF}     & \uline{23.73} & \uline{92.9} & \uline{92.9} & \uline{92.9} & \uline{92.8} \\
      \midrule

      % ----------------- I3D -----------------
      \multirow{5}{*}{I3D}
        & RGB     & --      & 12.29 & 87.8 & 88.4 & 88.6 & 88.5 \\
        & \multirow{4}{*}{RGB+MV}
        & \textbf{Concat} & \textbf{12.36} & \textbf{99.0} & \textbf{99.0} & \textbf{99.0} & \textbf{99.0} \\
        &         & \uline{JE}      & \uline{27.27} & \uline{97.7} & \uline{97.8} & \uline{97.8} & \uline{97.8} \\
        &         & CA      & 12.34 & 95.2 & 95.7 & 95.9 & 95.7 \\
        &         & MAF     & 12.54 & 96.8 & 96.5 & 96.7 & 96.5 \\
      \midrule

      % ----------------- TSN -----------------
      \multirow{5}{*}{TSN}
        & RGB     & --      & 23.52 & 95.7 & 95.4 & 95.6 & 95.4 \\
        & \multirow{4}{*}{RGB+MV}
        & \textbf{Concat} & \textbf{23.53} & \textbf{98.2} & \textbf{98.2} & \textbf{98.2} & \textbf{98.2} \\
        &         & \uline{JE }     & \uline{48.13} & \uline{95.9} & \uline{96.3} & \uline{96.4} & \uline{96.3} \\
        &         & CA      & 23.53 & 83.9 & 85.3 & 84.1 & 84.1 \\
        &         & MAF     & 23.73 & 94.1 & 92.4 & 95.9 & 92.4 \\
      \bottomrule
    \end{tabular}%
    }
    \vspace{-4pt}
\end{table}

\noindent
\textbf{Motion Density Thresholds for MAF.}\
Let \(v_t\colon \Omega \to \mathbb{R}^2\) be the MV field on \(\Omega=\{1,\dots,H\}\times\{1,\dots,W\}\). Define the binary mask
\begin{equation}
M_t(x,y) = \mathbf{1}\{\lVert v_t(x,y)\rVert_2 \ge \tau\},
\label{eq:mask}
\end{equation}
with detector threshold \(\tau>0\). The motion density of frame \(t\) is
\begin{equation}
d_t = \frac{1}{HW}\sum_{(x,y)\in\Omega} M_t(x,y) \in [0,1].
\label{eq:dt}
\end{equation}
\noindent
Let \(\alpha_{\ell}, \alpha_{h}\in(0,1)\) denote the quantile levels that parameterize the routing policy, with \(0<\alpha_{\ell}<\alpha_{h}<1\). Let \(\mathcal{D}=\{d_i\}_{i=1}^{N}\) be the set of framewise densities and let \(Q(\alpha)\) denote the empirical \(\alpha\)-quantile on \(\mathcal{D}\). The routing policy sets
\begin{equation}
p_{\text{low}}=\max\big(Q(\alpha_{\ell}),\,\varepsilon\big),
\qquad
p_{\text{high}}=Q(\alpha_{h}).
\label{eq:plowphigh}
\end{equation}
Here \(\varepsilon>0\) is a small constant to mitigate zero inflation when lower quantiles equal zero. 

\section{Results}
%We evaluate the effects of the integration of motion vectors as features in downstream classification tasks. In Table \ref{tab:dist_metrics}, the distance and divergence metrics between the motion vector distributions of synthetic video models and real video data is reported. Video generation models include t2v, cogvideo, vc2, svd, and pika. Kullback-Leibler divergences (KL) is computed with respect to both the real (P) and synthetic (Q) distributions, Jensen-Shannon divergence (JS), and Wasserstein Distance (WD). The distance is evaluated on two representations: motion vector sum, which quantifies the magnitude of motion vector displacement per frame and motion entropy, which captures the motion vector magnitude distributions. Lower scores indicate better results, indicating that pika achievest the best performance with the motion vector sum KL(P$\parallel$Q)=0.074 and JS=0.019. The performance across all models is illustrated in Figure \ref{fig:enter-label} where synthetic models are compared against real motion profiles. 

%In Table \ref{tab:class_results}, the difference in performance of RGB and RGB + MV is presented. ResNet Vision Transformer (ViT) across different depth and patch sizes are evaluated, showing that the addition of motion vectors as features consistently improves the performance of most architectures. Every model except ViT-B/32 has shown improved performance with the implementation of motion vector features.

We evaluate our method in two primary directions: (1) statistical comparisons of MV distributions between real and generated videos, and (2) classification performance when MVs are incorporated into standard video classifiers. These experiments assess both the fidelity of motion dynamics in synthetic videos and the utility of MVs as features for discriminative tasks. Our experiments use a subset of the GenVidBench dataset \cite{ni2025genvidbenchchallengingbenchmarkdetecting}, which is a large-scale benchmark designed to detect AI-generated videos. The dataset contains videos generated by several video generators and real (non-synthetic) videos drawn from multiple sources. In total, the dataset covers rich semantic content and supports cross-source evaluation.

\subsection{Statistical Alignment Analysis of Motion Vectors}

\begin{figure}[!htb]
    \centering
    \includegraphics[width=1\linewidth]{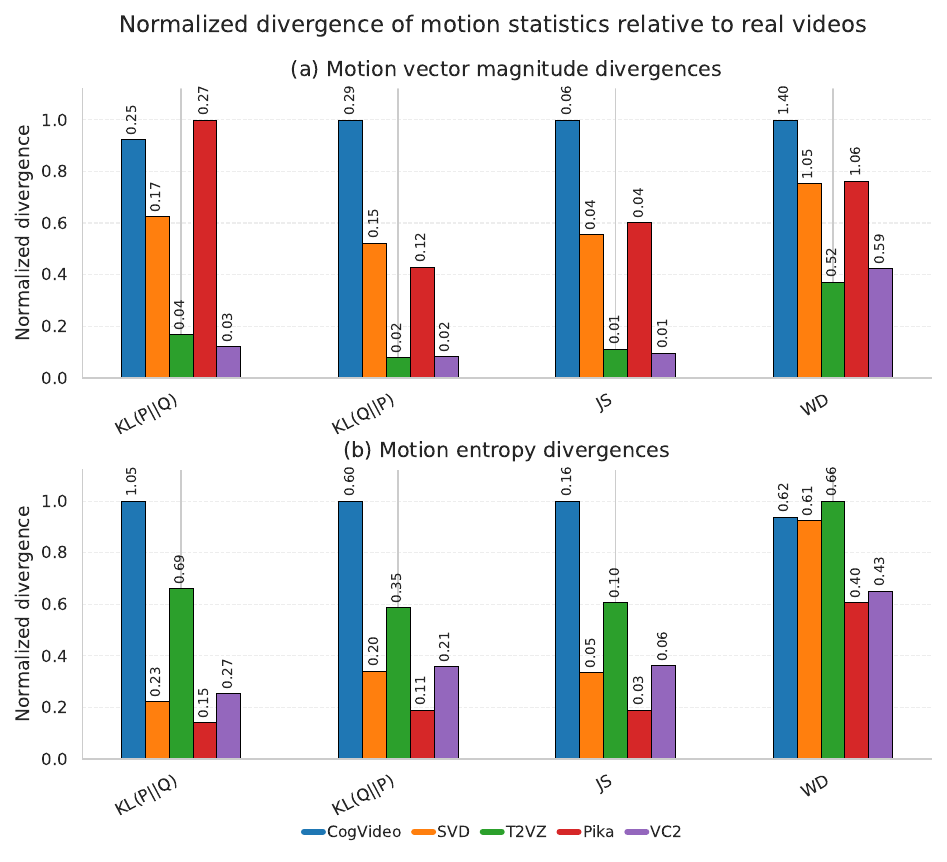}
  \caption{Normalized divergences of motion statistics for all models relative to real videos. 
Panel (a) reports divergences computed from motion vector magnitude metrics and panel (b) shows divergences based on motion entropy metrics. 
For each model, values are normalized per metric to highlight relative discrepancies across motion characteristics.  Lower values indicate closer alignment with real video dynamics, while higher values reflect stronger deviations in temporal realism. Pika aligns best with real motion entropy; VC2 and T2VZ achieve lowest divergences for motion-vector sums. Full results are in Appendix \ref{subsec:distmet}. }
  \label{fig:dist_metrics}
\end{figure}

Figure~\ref{fig:dist_metrics} summarizes the divergences between real and generated motion statistics, while Figure~\ref{fig:dist} provides visual comparisons of their empirical distributions. For motion entropy, Pika exhibits the strongest alignment with real videos, consistently yielding the lowest divergences across all metrics, with SVD following closely as the second-best performer. In contrast, MV sums reveal a more fragmented landscape: VC2 and T2VZ achieve the best scores on most measures, indicating that they capture aggregate displacement more effectively than other models. CogVideo stands out as the weakest across both statistics, likely due to its low frame rate and hierarchical generation strategy, which limit its ability to sustain smooth and coherent motion. The distribution plots in Figure~\ref{fig:dist} ($\textsc{top}$) reinforce these findings. Pika tracks the real entropy distribution most closely, both in average levels and in long-tail behavior, while SVD remains competitive across most regions. VC2 and T2VZ align more strongly with the real distributions of MV sums, consistent with their favorable scores in Figure~\ref{fig:dist_metrics}, although their entropy traces deviate more noticeably. In Figure~\ref{fig:dist} ($\textsc{top}$) All models exhibit oscillatory patterns in the per-frame averages, which stem from interpolation effects rather than intrinsic model behavior. The  Figure~\ref{fig:dist} ($\textsc{middle}$) row reveals that generator flow fields are sparser and more piecewise constant, with short streaks and directional clustering that are less common in real footage. Figure~\ref{fig:dist} ($\textsc{bottom}$) shows class-conditional magnitude heatmaps in which synthetic models exhibit center bias and other artifacts, while real videos distribute motion more diffusely across the frame. Taken together, results indicate that tested generators capture coarse motion but struggle to maintain temporally coherent micro-dynamics and spatially uniform motion allocation. 
\vspace{-0.3em}
\subsection{Classification Results}
% Performance with Motion Vectors

To assess the discriminative value of MVs, we train several deep classifiers using either RGB frames alone or a combined RGB and MV input representation. Table~\ref{tab:rgbmv-singlecol} reports performance metrics (including accuracy, F1 score, precision, and recall) for three ResNet \cite{resnet} models, I3D \cite{i3d}  and TSN \cite{tsn} backbones across both input settings. To assess the discriminative value of MVs, we train several deep classifiers using either RGB frames alone or a combined RGB and MV input representation. Table~\ref{tab:rgbmv-singlecol} reports performance metrics for all tested backbones across both input settings.

The results indicate a consistent performance boost when incorporating MVs alongside RGB inputs. Across all architectures, the fusion of motion and appearance cues enhances temporal discriminability and stabilizes per-class predictions. Among the ResNet variants, ResNet-34 and ResNet-18 demonstrate substantial improvements under joint embedding (JE) and cross-attention (CA) schemes, achieving up to 97.4\% accuracy. In contrast, ResNet-50 shows relatively smaller gains, likely due to its stronger representational capacity already modeling partial motion through stacked receptive fields. ResNet-50 also demonstrates a decline in performance compared to its smaller variants due to overfitting and redundancy between spatial and motion representations. For spatiotemporal architectures such as I3D, integrating codec-level MVs yields striking improvements even with simple channel concatenation, reaching 99.0\% accuracy. This observation highlights the complementary role of MVs to optical-flow-like patterns learned within 3D convolutions. Finally, TSN exhibits notable enhancement under concatenation and MAF fusion, confirming that even sparse temporal samplers can leverage coarse motion priors to improve segment-level consistency. Overall, learned fusion approaches such as JE and CA outperform naïve concatenation, validating that structured alignment between modalities is essential. The proposed MAF further provides a balance between interpretability and performance by adaptively gating motion contributions based on motion density, achieving competitive accuracy across all backbones. It is also worth noting that the JE strategy, while achieving the highest accuracy across several models, approximately doubles the number of trainable parameters due to its use of parallel embedding branches for RGB and motion features. This added capacity likely contributes to the improved performance but comes at a cost of both increased model complexity and training time.

\begin{table}[t]
  \centering
  \small
  \setlength{\tabcolsep}{3pt}
  \renewcommand{\arraystretch}{0.96}
  \caption{\textbf{Impact of Motion Vectors (MV) on Synthetic Video Classification.}
Comparison of ResNet performance using RGB vs.\ RGB + MV inputs with concatenation fusion across generation models. Values in \%. $\uparrow$ indicates improvement with MV.}
  \label{tab:class_vgModels_revised}
  \resizebox{\columnwidth}{!}{%
  \begin{tabular}{@{}l
      cc@{\hskip 4pt}cc@{\hskip 4pt}cc@{\hskip 4pt}cc@{}}
    \toprule
    \multirow{2}{*}{\textbf{Model}} &
    \multicolumn{2}{c}{\textbf{Acc}} &
    \multicolumn{2}{c}{\textbf{F1}} &
    \multicolumn{2}{c}{\textbf{Prec}} &
    \multicolumn{2}{c}{\textbf{Rec}} \\
    \cmidrule(lr){2-3}\cmidrule(lr){4-5}\cmidrule(lr){6-7}\cmidrule(lr){8-9}
     & RGB & RGB+MV & RGB & RGB+MV & RGB & RGB+MV & RGB & RGB+MV \\
    \midrule
    \textsc{CogVideo} & 97.8 & \textbf{98.6}$^{\uparrow0.8}$ &
                        98.8 & \textbf{99.1}$^{\uparrow0.3}$ &
                        98.8 & \textbf{99.1}$^{\uparrow0.3}$ &
                        98.8 & \textbf{99.1}$^{\uparrow0.3}$ \\

    \textsc{Pika} & 99.5 & \textbf{99.8}$^{\uparrow0.3}$ &
                   99.8 & \textbf{99.9}$^{\uparrow0.1}$ &
                   99.8 & \textbf{99.9}$^{\uparrow0.1}$ &
                   99.8 & \textbf{99.9}$^{\uparrow0.1}$ \\

    \textsc{T2VZ} & \textbf{94.1} & 90.5$^{\downarrow3.6}$ &
                   \textbf{96.3} & 92.3$^{\downarrow4.0}$ &
                   \textbf{96.3} & 93.2$^{\downarrow3.1}$ &
                   \textbf{96.4} & 91.9$^{\downarrow4.5}$ \\

    \textsc{VC2} & 95.5 & \textbf{95.6}$^{\uparrow0.1}$ &
                  97.3 & \textbf{97.5}$^{\uparrow0.2}$ &
                  97.3 & \textbf{97.5}$^{\uparrow0.2}$ &
                  97.3 & \textbf{97.5}$^{\uparrow0.2}$ \\

    \textsc{SVD} & 96.9 & \textbf{97.7}$^{\uparrow0.8}$ &
                  98.6 & \textbf{98.9}$^{\uparrow0.3}$ &
                  98.6 & \textbf{98.9}$^{\uparrow0.3}$ &
                  \textbf{98.8} & 98.7$^{\downarrow0.1}$ \\
    \bottomrule
  \end{tabular}%
  }
  \vspace{-4pt}
\end{table}

\noindent
We further investigate model behavior by training ResNet classifiers separately on videos generated by each synthetic model. The results, presented in Table~\ref{tab:class_vgModels_revised}, show that when trained on Pika or CogVideo videos, classifiers achieve near-perfect performance, with Pika in particular reaching 0.998 accuracy and 0.999 F1, precision, and recall when MVs are included. Such results indicate that these models produce motion statistics that are highly distinguishable from real videos, meaning that despite strong perceptual quality, their temporal dynamics deviate significantly from natural motion. By contrast, models such as SVD and VC2 yield lower classification scores, even when augmented with MVs, suggesting that the motion they generate more closely aligns with real dynamics and is therefore harder for classifiers to separate. T2VZ represents an exception, as its performance decreases when MVs are included (0.905 with RGB+MV vs. 0.941 with RGB alone). This drop reflects the relatively sparse motion encoding exhibited by T2VZ, as also seen in its motion statistics. The limited temporal activity provides a weak signal, which, when fused with high-dimensional RGB inputs, may dilute discriminative features rather than enhance them. In this setting, the curse of dimensionality reduces classifier effectiveness, leading to degraded results. Together, these findings demonstrate that while MVs generally improve discriminability, their benefit depends on the richness of the underlying temporal patterns produced by the generative model.

\section{Conclusion}
We introduced a compressed-domain framework for evaluating and enhancing temporal fidelity in generative videos through MVs extracted directly from codec streams. By quantifying the distributional alignment between real and synthesized motion statistics, our approach exposes temporal inconsistencies often overlooked by frame-based metrics. The experiments demonstrate that video generators, while visually convincing, still diverge from real videos in fine-grained motion coherence. Moreover, integrating MVs with RGB features consistently improves discriminative performance across convolutional and transformer backbones, confirming that codec-level motion cues serve as lightweight yet informative temporal priors.
Overall, these findings suggest that compressed motion information offers a practical and scalable route toward temporally aware evaluation and learning. Future work will explore extending this analysis to generative adversarial objectives, developing unified motion-perceptual metrics, and leveraging MVs for efficient temporal regularization during video synthesis.

% references without citing itry in the main text, use \nocite

{
    \small
    \bibliographystyle{ieeenat_fullname}
    \bibliography{main}              
}

%%%%%%%%%%%%%%%%%%%%%%%%%%%%%%%%%%%%%%%%%%%%%%%%%%%%%%%%%%%%%%%%%%%%%%%%%%%%%%%
%%%%%%%%%%%%%%%%%%%%%%%%%%%%%%%%%%%%%%%%%%%%%%%%%%%%%%%%%%%%%%%%%%%%%%%%%%%%%%%
% APPENDIX
%%%%%%%%%%%%%%%%%%%%%%%%%%%%%%%%%%%%%%%%%%%%%%%%%%%%%%%%%%%%%%%%%%%%%%%%%%%%%%%
%%%%%%%%%%%%%%%%%%%%%%%%%%%%%%%%%%%%%%%%%%%%%%%%%%%%%%%%%%%%%%%%%%%%%%%%%%%%%%%
\clearpage
\appendix
\section{Spatial Analysis}
\subsection{Direction Distribution}
\label{subsec:direction_dist}
For each motion-vector field
\begin{equation}
V_t(u,v) = \big(V^x_t(u,v), V^y_t(u,v)\big)
\end{equation}
, we compute the angular orientation 
\begin{equation}
\theta_t(u,v) = \operatorname{atan2}\!\big(V^y_t(u,v), V^x_t(u,v)\big),
\end{equation}

which assigns to every spatial coordinate \((u,v)\) and frame \(t\) a direction on the unit circle. Aggregating \(\theta_t(u,v)\) across all clips belonging to class \(c\) yields an empirical angular density \(f_c(\theta)\). We visualize this density via rose diagrams, where bar heights are proportional to \(f_c(\theta)\); directional biases or isotropy are readily apparent when comparing classes. Figure~\ref{fig:direction-distribution} contrasts these distributions, highlighting the degree to which each model adheres to or diverges from uniform motion directionality.

\begin{figure*}
    \centering
    \includegraphics[width=0.75\linewidth]{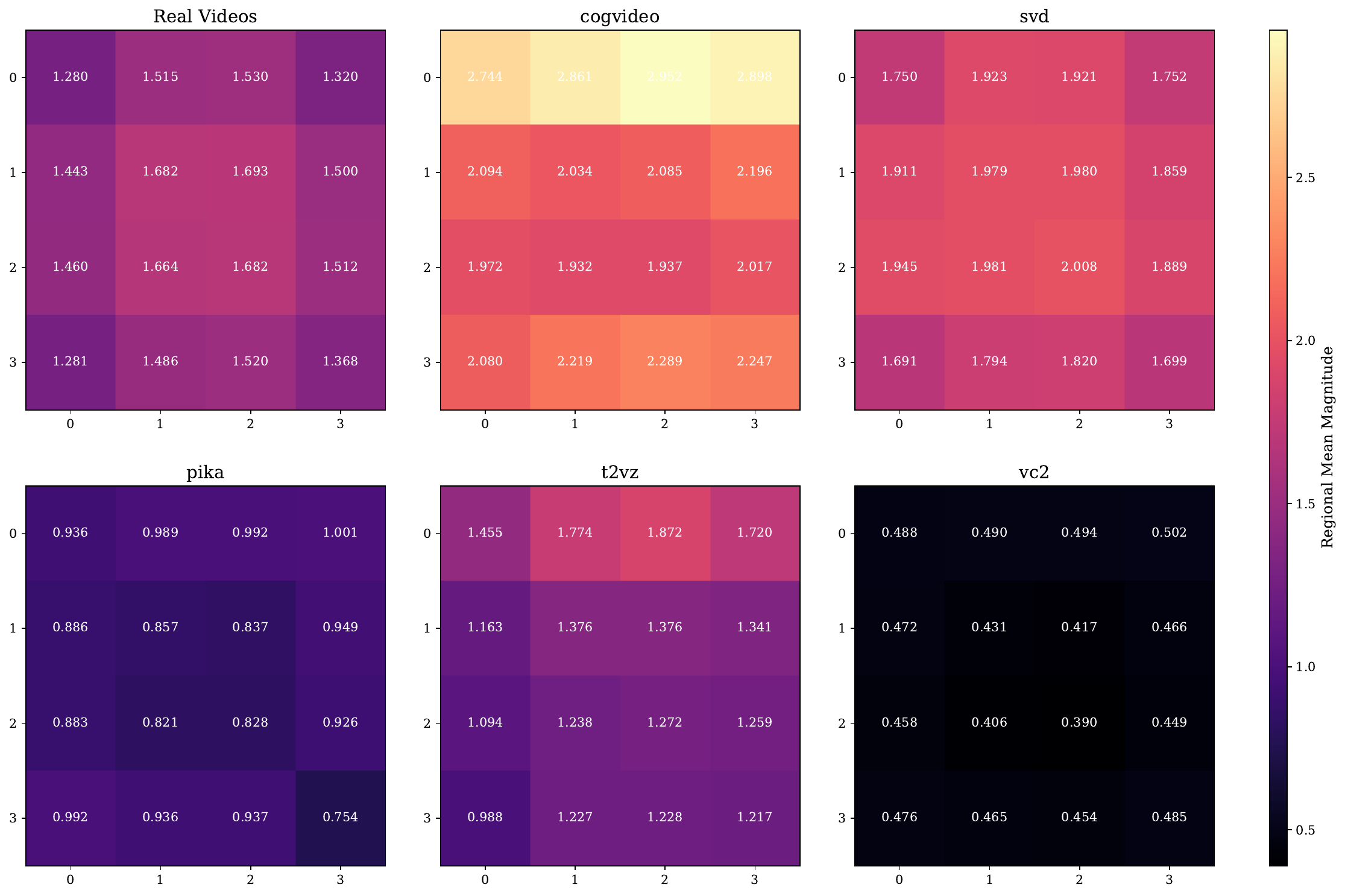}
    \caption{Regional motion-energy profiles on a \(4\times4\) grid. Each cell encodes the expected per-region magnitude \(\bar{s}_c(r,c)\) (and/or its normalized share \(p_c(r,c)\)), averaged over frames and clips within each class; higher values indicate regions where motion energy concentrates.}
    \label{fig:regional-patterns}
\end{figure*}

\subsection{Magnitude Heatmap}
\label{subsec:heatmap}
To localize motion energy across the image plane, we use the per-pixel Euclidean magnitude of the previously defined motion-vector field,
\begin{equation}
m_t(u,v) \;=\; \big\|V_t(u,v)\big\|_2 \;=\; \sqrt{\big(V^x_t(u,v)\big)^2 + \big(V^y_t(u,v)\big)^2}.
\end{equation}
After bilinearly resizing fields to a common grid \(\Omega \subset \{1,\dots,56\}\times\{1,\dots,56\}\), a clip \(i\) in class \(c\) with frame index set \(T_i\) is summarized by its time-averaged magnitude map
\begin{equation}
h_i(u,v) \;=\; \frac{1}{|T_i|}\sum_{t\in T_i} m^{(i)}_t(u,v).
\end{equation}
The class-conditional heatmap is the empirical mean across clips,
\begin{equation}
\bar{h}_c(u,v) \;=\; \frac{1}{N_c}\sum_{i\in \mathcal{D}_c} h_i(u,v),
\end{equation}
and is min–max normalized for visualization,
\begin{equation}
\tilde{h}_c(u,v) \;=\; \frac{\bar{h}_c(u,v) - \min_{(u',v')\in\Omega}\bar{h}_c(u',v')}{\max_{(u',v')\in\Omega}\bar{h}_c(u',v') - \min_{(u',v')\in\Omega}\bar{h}_c(u',v')}.
\end{equation}
Brighter regions in Figure~\ref{fig:dist} (\textsc{bottom}) mark spatial locations with higher expected motion energy, exposing center–edge biases and other class-specific regularities.

Table~\ref{tab:heatmap} presents per-class spatial statistics derived from average motion magnitude heatmaps. Each class is summarized using global statistics (mean, standard deviation, maximum, and minimum), spatial variance, peak-to-mean ratio, and region-specific metrics computed from central and peripheral areas. The central region corresponds to the middle 50\% of pixels, while the edge region is defined as a 10\% thick outer band along the image borders.

To summarize the coarse spatial allocation of motion energy across the frame (Figure~\ref{fig:regional-patterns}), we partition the common grid \(\Omega\) into a \(4\times4\) regular lattice \(\mathcal{P}=\{R_{r,c}\}_{r,c=1}^{4}\) of equal-area regions. For each frame \(t\) and region \(R_{r,c}\), we compute the region-mean magnitude
\begin{equation}
s_t(r,c) \;=\; \frac{1}{|R_{r,c}|}\sum_{(u,v)\in R_{r,c}} m_t(u,v),
\end{equation}
and aggregate over a clip \(i\) with frame set \(T_i\) as
\begin{equation}
s_i(r,c) \;=\; \frac{1}{|T_i|}\sum_{t\in T_i} s^{(i)}_t(r,c).
\end{equation}
The class-conditional regional profile is the empirical mean across clips in class \(c\),
\begin{equation}
\bar{s}_c(r,c) \;=\; \frac{1}{N_c}\sum_{i\in\mathcal{D}_c} s_i(r,c),
\end{equation}
which we optionally convert to a composition-normalized share,
\begin{equation}
p_c(r,c) \;=\; \frac{\bar{s}_c(r,c)}{\sum_{r'=1}^{4}\sum_{c'=1}^{4}\bar{s}_c(r',c')}.
\end{equation}
As illustrated in Figure~\ref{fig:regional-patterns}, visualizing \(\bar{s}_c\) (and/or \(p_c\)) on the \(4\times4\) grid highlights center–edge biases, horizon alignment, and quadrant asymmetries that are less apparent at full spatial resolution.

\subsection{Flow Fields}
\label{subsec:flowfields}
To expose the instantaneous spatial organization of motion, we visualize representative flow fields for each class as illustrated in Figure \ref{fig:dist} (\textsc{MIDDLE}). For a clip \(i\) in class \(c\), let the per-frame energy be
\begin{equation}
e_i(t) \;=\; \mathbb{E}_{(u,v)\in\Omega}\big[m^{(i)}_t(u,v)\big],
\end{equation}
and choose a typical frame
\begin{equation}
t_i^\star \;=\; \arg\min_{t}\,\big|\,e_i(t) - \operatorname{median}_{t'} e_i(t')\,\big|.
\end{equation}
The selected field is \(F^{(i)}_c(u,v) = V^{(i)}_{t_i^\star}(u,v)\), resized to the common grid \(\Omega\subset\{1,\dots,56\}\times\{1,\dots,56\}\). For readability, we downsample to a coarse lattice \(\Lambda\) via block averaging \(D:\Omega\!\to\!\Lambda\), and render glyphs whose orientation matches the local direction and whose length encodes relative magnitude:
\begin{equation}
G^{(i)}_c(p) \;=\; \frac{D\!\big(F^{(i)}_c\big)(p)}{\,\max_{q\in\Lambda}\|D(F^{(i)}_c)(q)\|_2 + \varepsilon\,}, 
\quad p\in\Lambda,\ \varepsilon>0.
\end{equation}
This depiction preserves directional structure while normalizing within-panel scale, making coherent flows, expansion/contraction, and center–edge asymmetries visually comparable across classes.

\begin{table*}[!htb]
  \centering
  \small
  \setlength{\tabcolsep}{4pt}
  \renewcommand{\arraystretch}{0.97}
  \caption{\textbf{Per-class motion statistics for real and generated videos.}
  \textbf{S. Var.}: Spatial Variance, \textbf{P./M.}: Peak-to-Mean Ratio, 
  \textbf{C.M.}: Center Mean, \textbf{C. Std}: Center Std., 
  \textbf{E. Mean}: Edge Mean, \textbf{E. Std}: Edge Std., 
  \textbf{C./E.}: Center-to-Edge Ratio.}
  \label{tab:heatmap}
  \resizebox{0.8\textwidth}{!}{%
  \begin{tabular}{
    @{}l
    S[table-format=1.3] S[table-format=1.3] S[table-format=1.3] S[table-format=1.3]
    S[table-format=1.3] S[table-format=1.3]
    S[table-format=1.3] S[table-format=1.3]
    S[table-format=1.3] S[table-format=1.3]
    S[table-format=1.3]@{}}
    \toprule
    \multirow{2}{*}{\textbf{Class}} &
      \multicolumn{4}{c}{\textbf{Basic Statistics}} &
      \multicolumn{2}{c}{\textbf{Spatial}} &
      \multicolumn{2}{c}{\textbf{Center}} &
      \multicolumn{2}{c}{\textbf{Edge}} &
      \multicolumn{1}{c}{\textbf{Ratio}} \\
    \cmidrule(lr){2-5}
    \cmidrule(lr){6-7}
    \cmidrule(lr){8-9}
    \cmidrule(lr){10-11}
    \cmidrule(lr){12-12}
    & \textbf{Mean} & \textbf{Std} & \textbf{Max} & \textbf{Min} 
    & \textbf{S. Var.} & \textbf{P./M.}
    & \textbf{C.M.} & \textbf{C. Std} 
    & \textbf{E. Mean} & \textbf{E. Std} 
    & \textbf{C./E.} \\ 
    \midrule
    Real Videos   & 1.496 & 0.230 & 1.872 & 0.578 & 0.053 & 1.251 & 1.680 & 0.055 & 1.202 & 0.198 & 1.398 \\
    \textsc{CogVideo} & 2.285 & 0.423 & 3.716 & 1.113 & 0.179 & 1.626 & 1.997 & 0.176 & 2.347 & 0.569 & 0.851 \\
    \textsc{SVD}      & 1.869 & 0.222 & 2.351 & 0.925 & 0.049 & 1.258 & 1.987 & 0.086 & 1.613 & 0.233 & 1.232 \\
    \textsc{Pika}     & 0.908 & 0.126 & 1.208 & 0.009 & 0.016 & 1.331 & 0.836 & 0.041 & 0.906 & 0.218 & 0.923 \\
    \textsc{T2VZ}     & 1.350 & 0.295 & 2.668 & 0.497 & 0.087 & 1.976 & 1.316 & 0.109 & 1.313 & 0.435 & 1.002 \\
    \textsc{VC2}      & 0.459 & 0.045 & 0.612 & 0.291 & 0.002 & 1.334 & 0.411 & 0.024 & 0.492 & 0.042 & 0.836 \\
    \bottomrule
  \end{tabular}
  }
\end{table*}

\section{Statistical Analysis}
\begin{figure*}[t]
    \centering
    \begin{subfigure}{.49\linewidth}
        \centering
        \includegraphics[width=\linewidth]{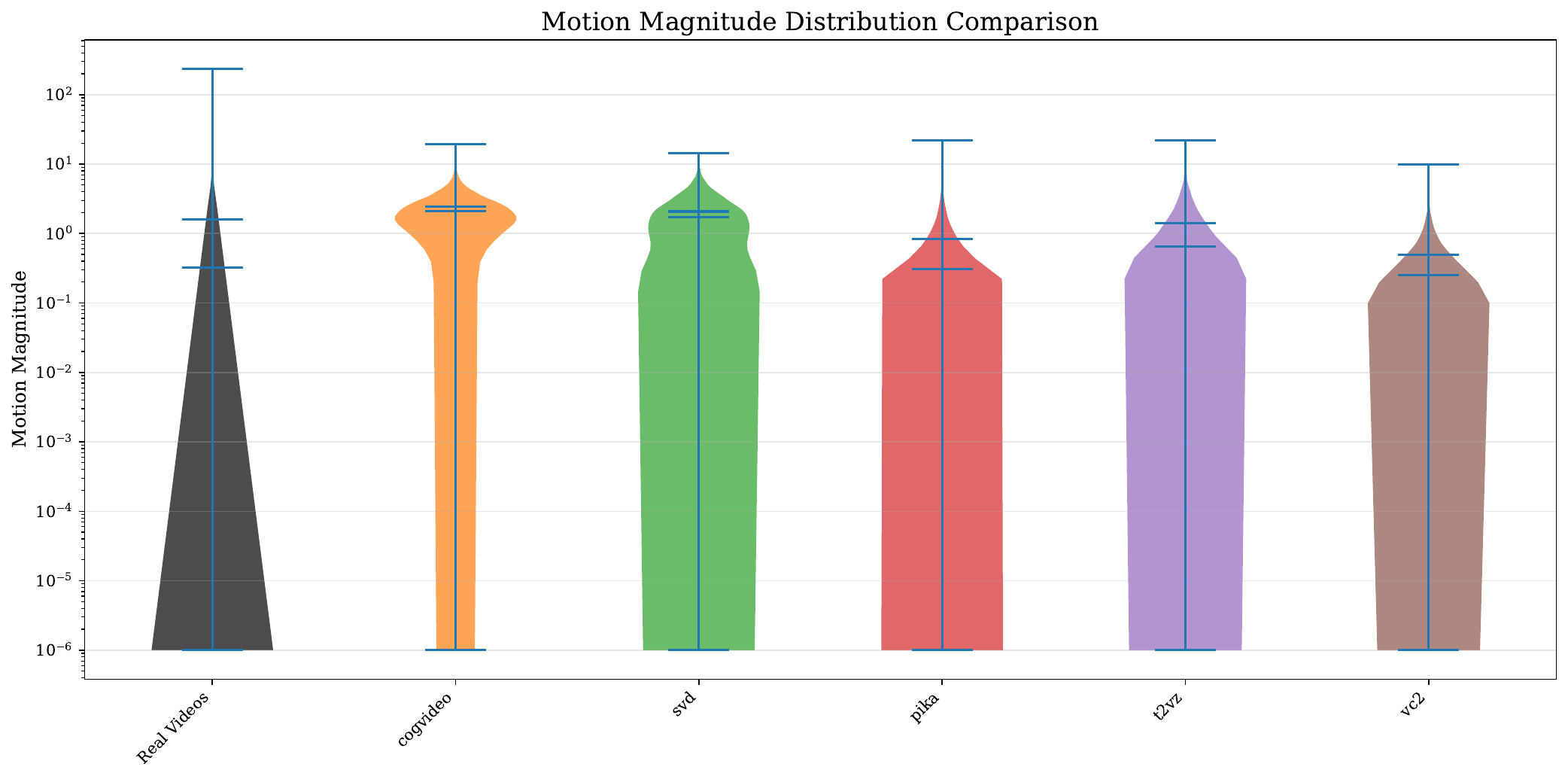}
        \caption{Magnitude distribution \(\{M_i\}\) (log-scale).}
        \label{fig:dist-violin-mag}
    \end{subfigure}
    \hfill
    \begin{subfigure}{.49\linewidth}
        \centering
        \includegraphics[width=\linewidth]{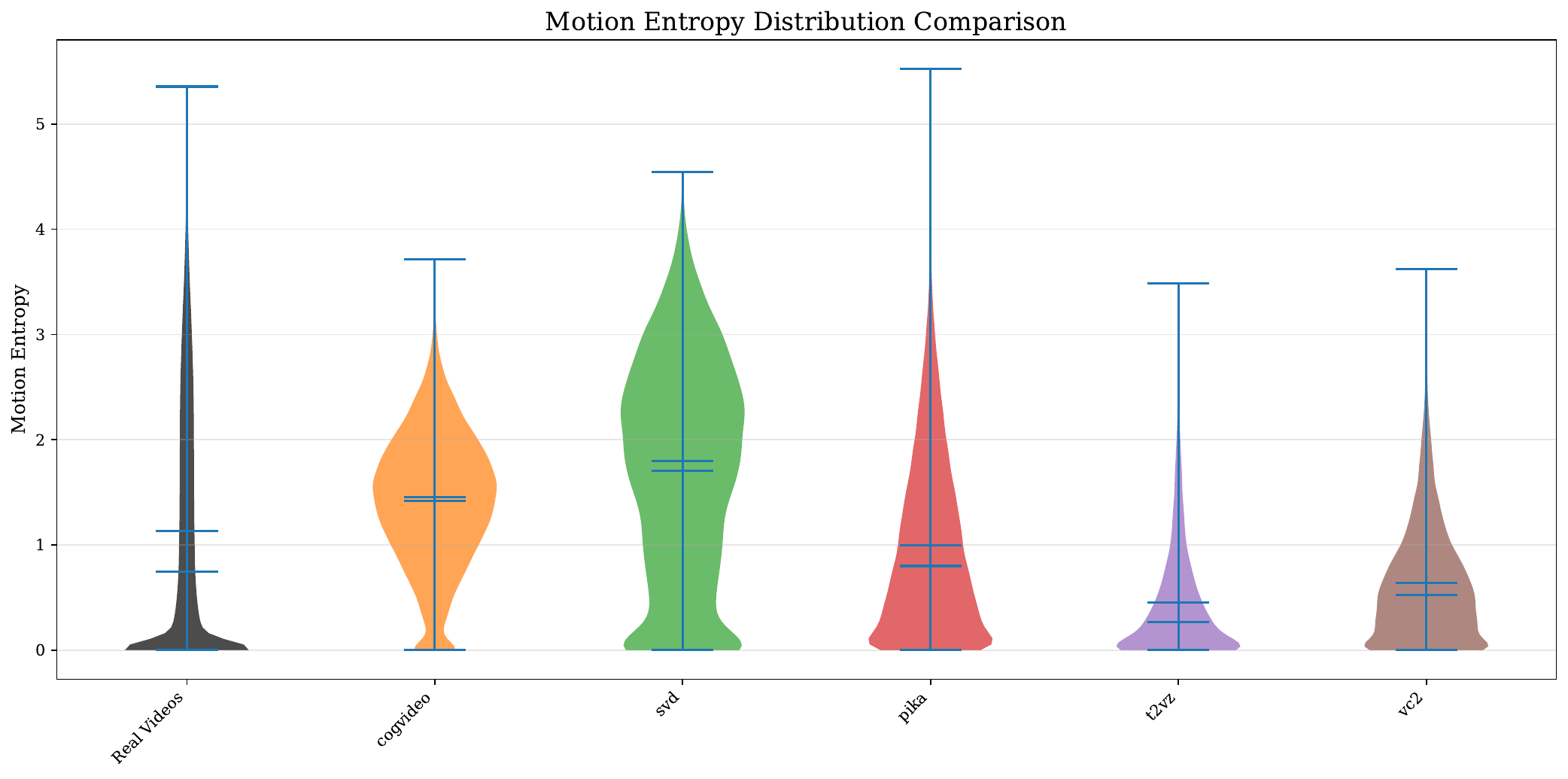}
        \caption{Entropy distribution \(\{H_i\}\).}
        \label{fig:dist-violin-ent}
    \end{subfigure}
    \caption{Class-conditional distributions of clip-level motion descriptors. (a) Mean motion magnitude \(M_i\) per clip, displayed on a log scale. (b) Motion entropy \(H_i\) per clip, averaging per-frame Shannon entropy of the magnitude field. Together, these summarize the strength and spatial complexity of motion across classes.}
    \label{fig:dist-violins}
\end{figure*}

To compare the intensity and spatial complexity of motion across classes (Figure~\ref{fig:dist-violins}, panels \subref{fig:dist-violin-mag}–\subref{fig:dist-violin-ent}), we summarize two clip-level descriptors derived from the magnitude field \(m_t(u,v)\). First, we compute the framewise mean magnitude
Use a line break with aligned equations:

\begin{equation}
\bar{m}*t = \mathbb{E}*{(u,v)\in\Omega}!\big[m_t(u,v)\big],
\end{equation}

\begin{equation}
M_i = \frac{1}{|T_i|}\sum_{t\in T_i} \bar{m}^{(i)}_t
\end{equation}
yielding per-class distributions \(\{M_i : i\in\mathcal{D}_c\}\). Second, we quantify spatial complexity via the Shannon entropy of the discretized magnitude field: with bins \(\{I_k\}_{k=1}^K\),
\begin{align}
p_t(k) &= \frac{1}{|\Omega|}\sum_{(u,v)\in\Omega}\mathbf{1}\{m_t(u,v)\in I_k\}, \\[4pt]
H_t &= -\sum_{k=1}^K p_t(k)\log p_t(k), \\[4pt]
H_i &= \frac{1}{|T_i|}\sum_{t\in T_i} H_t.
\end{align}

Violin plots visualize the class-conditional distributions of \(\{M_i\}\) and \(\{H_i\}\): lower \(H_i\) indicates concentrated (structured) motion energy, while higher \(H_i\) reflects diffuse motion; magnitude violins are shown on a logarithmic scale to accommodate heavy-tailed variability.

\begin{figure}
\includegraphics[width=\linewidth]{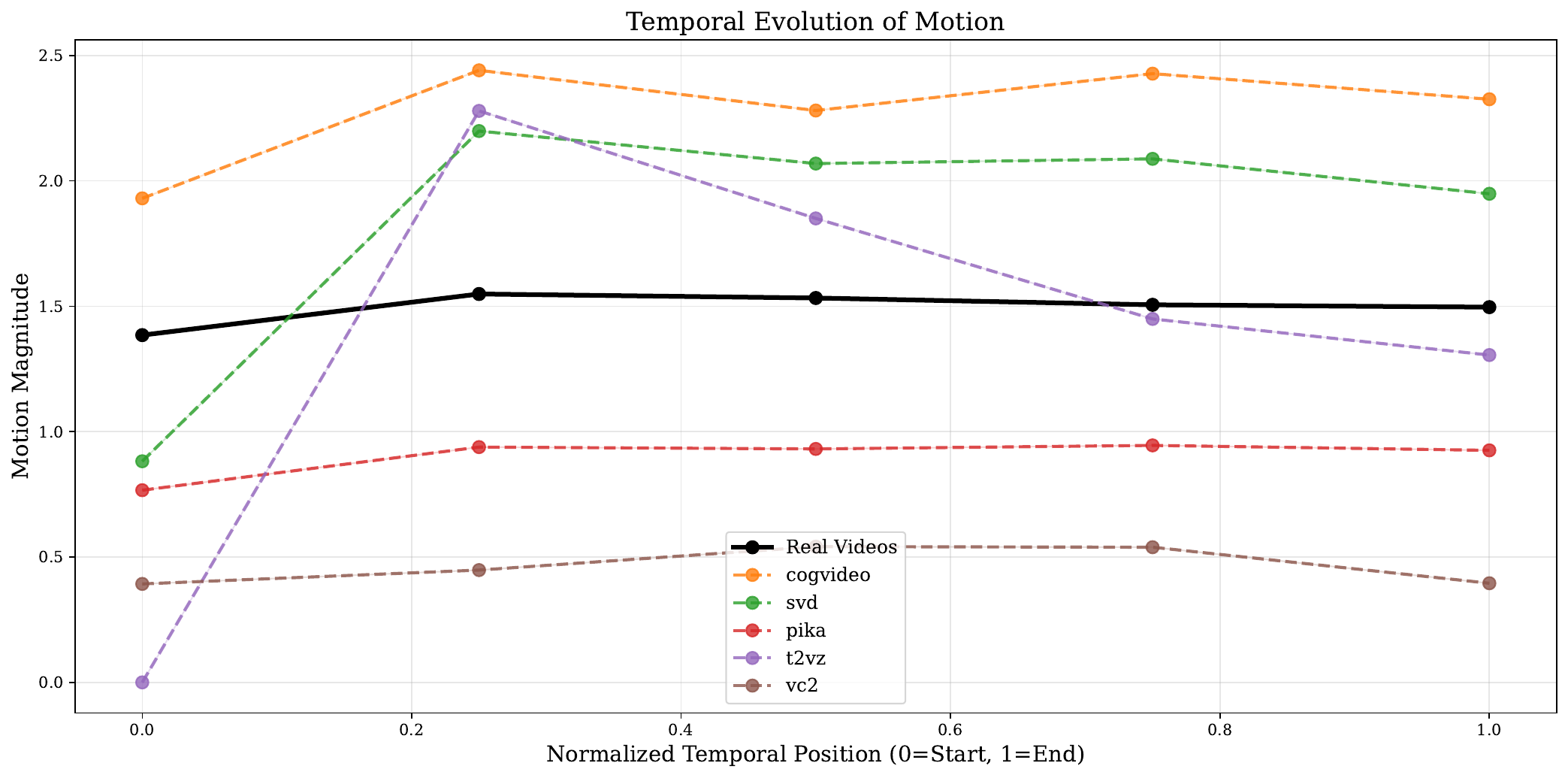}
    \caption{Temporal evolution of motion magnitude by class. Curves show the class-mean segment-wise magnitude \(\mu_c(j)\) over \(n=5\) equal-duration segments (relative time \(\tau\in[0,1]\)). Higher trajectories reflect stronger motion; increasing or decreasing trends reveal onset/decay dynamics.}
    \label{fig:motion-evolution}
\end{figure}

\subsection{Motion Evolution}
To characterize how motion intensity evolves over the duration of each clip (Figure~\ref{fig:motion-evolution}), we partition the frame index set \(T_i\) into \(n\) equal-length segments \(S_{i,1:n}\) (here \(n=5\)). Using the per-frame mean magnitude of the motion field,
\begin{equation}
\bar{m}_t \;=\; \mathbb{E}_{(u,v)\in\Omega}\!\big[m_t(u,v)\big],
\end{equation}

we define the segment-wise descriptor
\begin{equation}
M_{i,j} \;=\; \frac{1}{|S_{i,j}|}\sum_{t\in S_{i,j}} \bar{m}^{(i)}_t,\qquad j=1,\dots,n.
\end{equation}
Class-conditional temporal trajectories are obtained by averaging across clips,
\begin{equation}
\mu_c(j) \;=\; \frac{1}{N_c}\sum_{i\in\mathcal{D}_c} M_{i,j},
\end{equation}
with variability summarized via pointwise standard errors (and 95\% bootstrap confidence intervals when enabled). We report time in relative units \(\tau_j = \frac{j-1}{n-1}\in[0,1]\) to make clips of different lengths comparable; rising or decaying profiles of \(\mu_c(j)\) indicate systematic onset/offset patterns in motion strength.

\vspace{-0.4em}
\subsection{Distribution Metrics}
Distribution metrics of real and generated video motion statistics are reported in Table \ref{tab:dist_metrics}. KL, JS and Wasserstein divergences over MV sums and motion entropy for six generative models. 
\label{subsec:distmet}

\begin{table}[!htb]
  \centering
  \small
  \setlength{\tabcolsep}{3pt}
  \renewcommand{\arraystretch}{0.97}
  \caption{\textbf{Pika aligns best with real motion entropy; VC2 and T2VZ achieve lowest divergences for motion-vector sums.}
  Divergences to real video motion statistics (lower is better). \textbf{Bold} = best, \uline{underline} = second-best.}
  \label{tab:dist_metrics}
  \resizebox{1\columnwidth}{!}{%
  \begin{tabular}{
    @{}l
    S[table-format=1.3] S[table-format=1.3] S[table-format=1.3] S[table-format=1.3]
    S[table-format=1.3] S[table-format=1.3] S[table-format=1.3] S[table-format=1.3]@{}}
    \toprule
    \multirow{2}{*}{\textbf{Model}} &
      \multicolumn{4}{c}{\textbf{Motion Vector Sum}} &
      \multicolumn{4}{c}{\textbf{Motion Entropy}} \\
    \cmidrule(lr){2-5} \cmidrule(lr){6-9}
    & \textbf{KL($P\!\parallel\!Q$)} & \textbf{KL($Q\!\parallel\!P$)} & \textbf{JS} & \textbf{WD}
    & \textbf{KL($P\!\parallel\!Q$)} & \textbf{KL($Q\!\parallel\!P$)} & \textbf{JS} & \textbf{WD} \\
    \midrule
    \textsc{CogVideo} & 0.247 & 0.289 & 0.063 & 1.395 & 1.048 & 0.598 & 0.158 & 0.616 \\
    \textsc{SVD}      & 0.167 & 0.151 & 0.035 & 1.050 & \uline{0.234} & \uline{0.204} & \uline{0.053} & 0.607 \\
    \textsc{T2VZ}    & \uline{0.045} & \textbf{0.023} & \uline{0.007} & \textbf{0.518} & 0.691 & 0.351 & 0.096 & 0.657 \\
    \textsc{Pika}     & 0.267 & 0.124 & 0.038 & 1.065 & \textbf{0.147} & \textbf{0.113} & \textbf{0.030} & \textbf{0.399} \\
    \textsc{VC2}      & \textbf{0.032} & \uline{0.024} & \textbf{0.006} & \uline{0.591} & 0.266 & 0.214 & 0.057 & \uline{0.428} \\
    \bottomrule
  \end{tabular}
  }
  \vspace{-4pt}
\end{table}

\section{Experimental Details}
\vspace{-0.4em}

All experiments are conducted on a single NVIDIA A100 Tensor Core GPU for 10 epochs. We set the mask quantile to \(q_{\mathrm{mv}}\), and use \(p_{\text{low}}=0.0001\) and \(p_{\text{high}}=0.4949\). These thresholds yield a routing distribution of 37.3\% for the low route (skip MV), 32.8\% for the mid route (light fusion), and 29.9\% for the high route (heavy fusion). We built our implementation on top of the Motion Vector Extractor library \cite{MV_Extractor}, available at: \url{https://github.com/LukasBommes/mv-extractor}

%%%%%%%%%%%%%%%%%%%%%%%%%%%%%%%%%%%%%%%%%%%%%%%%%%%%%%%%%%%%%%%%%%%%%%%%%%%%%%%
%%%%%%%%%%%%%%%%%%%%%%%%%%%%%%%%%%%%%%%%%%%%%%%%%%%%%%%%%%%%%%%%%%%%%%%%%%%%%%%

\end{document}